\documentclass[10pt,twocolumn,letterpaper]{article}

\usepackage[pagenumbers]{iccv} 

\definecolor{cvprblue}{rgb}{0.21,0.49,0.74}
\usepackage[pagebackref,breaklinks,colorlinks,allcolors=cvprblue]{hyperref}
\usepackage{multirow}
\usepackage{booktabs}
\usepackage{makecell}
\usepackage{array}
\usepackage{amssymb}
\usepackage{colortbl}
\usepackage{graphicx}
\usepackage{caption}

\newcommand{\blfootnote}[1]{%
  \begingroup
  \renewcommand\thefootnote{}
  \footnotetext{#1}%
  \endgroup
}

\def\paperID{4646} 
\def\confName{ICCV}
\def\confYear{2025}

\title{Reducing CT Metal Artifacts 
by Learning Latent Space Alignment \\ with Gemstone Spectral Imaging Data}

\author{
Wencheng Han$^{1*}$,
Dongqian Guo$^{1*}$,
Xiao Chen$^{2}$,
Pang Lyu$^{3}$,
Yi Jin$^{2}$,
Jianbing Shen$^{1\dagger}$\\
$^1$SKL-IOTSC, CIS, University of Macau\\
$^2$Department of Orthopedics, People's Hospital of Zhengzhou University,\\ Henan Provincial People's Hospital\\
$^3$Zhongshan Hospital, Fudan University\\
 {\tt\small wencheng256@gmail.com, jianbingshen@um.edu.mo}\\
}

\begin{document}
\maketitle

\begin{abstract}

Metal artifacts in CT slices have long posed challenges in medical diagnostics. These artifacts degrade image quality, resulting in suboptimal visualization and complicating the accurate interpretation of tissues adjacent to metal implants. To address these issues, we introduce the Latent Gemstone Spectral Imaging (GSI) Alignment Framework, which effectively reduces metal artifacts while avoiding the introduction of noise information. Our work is based on a key finding that even artifact-affected ordinary CT sequences contain sufficient information to discern detailed structures. The challenge lies in the inability to clearly represent this information.
To address this issue, we developed an Alignment Framework that adjusts the representation of ordinary CT images to match GSI CT sequences. GSI is an advanced imaging technique using multiple energy levels to mitigate artifacts caused by metal implants. By aligning the representation to GSI data, we can effectively suppress metal artifacts while clearly revealing detailed structure, without introducing extraneous information into CT sequences. To facilitate the application, we propose a new dataset, Artifacts-GSI, captured from real patients with metal implants, and establish a new benchmark based on this dataset. Experimental results show that our method significantly reduces metal artifacts and greatly enhances the readability of CT slices. \textit{All our code and data are available at: \url{https://um-lab.github.io/GSI-MAR/}}
\end{abstract} 
\blfootnote{$*$Equal contribution. $\dagger$Corresponding author: \textit{Jianbing Shen}.}

\section{Introduction}

\begin{figure*}[ht]
    \centering
    \includegraphics[width=1.0\textwidth]{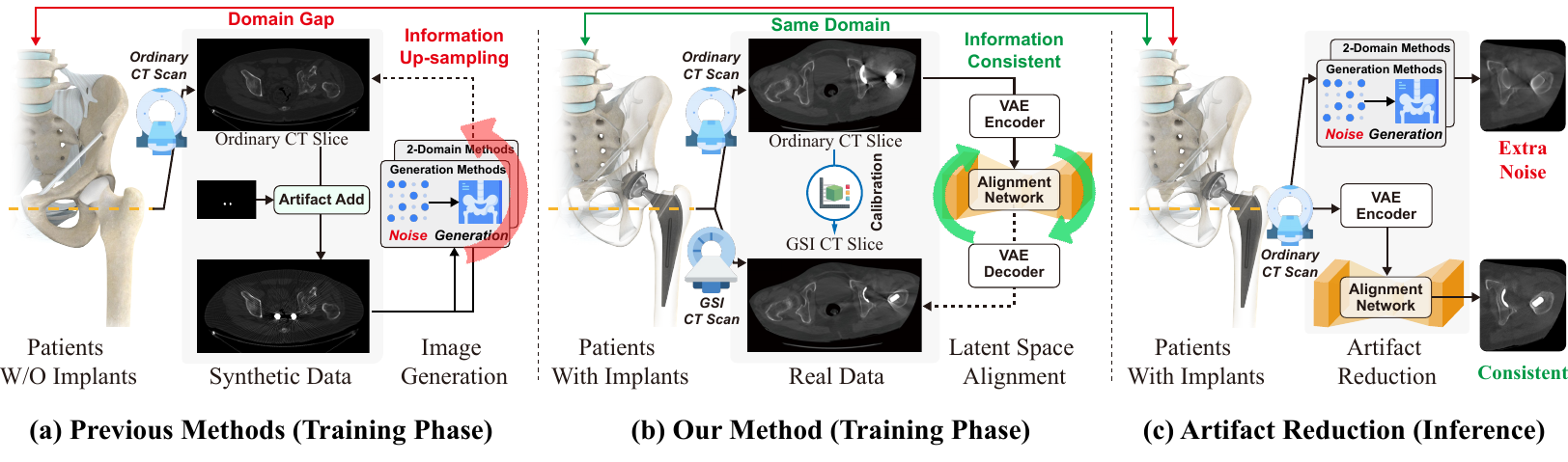}
    \vspace{-4mm}
    \caption{\textbf{Comparison of Artifacts Reduction Pipelines}. \textbf{(a)} Most previous methods rely on synthetic artifact data derived from clean CT sequences of patients without implants. 
    Additionally, many methods use image generation algorithms, which may introduce extraneous information, potentially compromising the reliability of the resulting CT sequences.
    \textbf{(b)} In contrast, our method utilizes real artifact CT pairs for training, effectively bridging the domain gap. Our approach employs a representation alignment algorithm, maintaining information consistency.
    \textbf{(c)} We provide a comparison of inference results between our method and previous methods to illustrate the effectiveness of our approach.
    }
    \label{fig:motivation}
    \vspace{-4mm}
\end{figure*}


Computed tomography (CT)~\cite{hounsfield1973computerized} plays a crucial role in radiology by offering detailed information for diagnosis and treatment planning. 
However, CT artifact from metal implants is a persistent challenge~\cite{de1998metal, selles2023advances}, particularly as the demand for implants rises due to global aging and a surge in trauma cases. These artifacts primarily arise from photon starvation and beam hardening~\cite{yadava2014reduction}. High atomic number materials in implants absorb excessive X-ray photons, leading to low photon counts at the detector and modified beam energies. 
Such artifacts degrade image quality, obscure details, and complicate the accurate assessment of tissues around metal implants~\cite{boas2012ct}. Moreover, they may introduce misleading image content, raising the risk of misdiagnosis. Therefore, developing novel methods to reduce metal artifacts while maintaining the authenticity and integrity of CT images is of great importance in clinical practice.

Manufacturers have attempted to reduce metal artifacts by improving the imaging techniques. 
For instance, dual-energy CT~\cite{johnson2007material, slavic2022dual} employs two
X-ray tubes and detectors operating at different voltages, while gemstone spectral imaging (GSI) CT~\cite{slavic2017gsi, lee2012metal} enables rapid energy switching of X-ray tubes during scans.
These methods improve image quality and diagnostic accuracy by capturing tissue attenuation information more effectively across different energy levels, thus naturally helping to suppress metal artifacts. 
However, these hardware-level upgrades are often costly and slow to iterate, resulting in poor accessibility in many medical facilities.
Benefiting from advancements in computer vision technology, {recent methods~\cite{huang2018metal, selles2023ai}} have turned into deep learning algorithms to tackle the metal artifact reduction task. {Some researchers~\cite{park2018ct, zhu2021completion}} have focused on image processing techniques, aiming to restore areas affected by metal implants on the sinogram to eliminate artifacts~\cite{kalender1987reduction, meyer2012frequency}. Additionally, dual-domain methods combine information from both the sinogram and image domains to improve the metal reduction performance~\cite{lin2019dudonet, lyu2020encoding}.



Despite achieving significant success, these methods face limitations that hinder their broad application, as shown in \cref{fig:motivation} (a) and (c). Firstly, most of these methods rely on synthetic data for training. These methods~\cite{zhang2018convolutional} add simulated artifacts to implant-free CT slices to create artifact-filled slices, while using the original clean slices as ground truth. However, synthetic data often fails to accurately represent real-world data distribution, creating a domain gap when applying these methods in clinical settings. Secondly, many existing approaches use image generation algorithms~\cite{liu2024unsupervised, wang2024metal, wang2018conditional} to recover missing information in artifact-affected areas. 
While, these methods risk introducing erroneous information, potentially compromising diagnostic accuracy.

To address these limitations and improve applicability in real medical scenarios, our work focuses on two key areas. Firstly, as shown in \cref{fig:motivation} (b), we introduce ArtifactGSI, a new metal artifacts reduction dataset collected from medical applications. This dataset includes CT scans from over 100 patients who have undergone joint arthroplasty surgery. Each patient's data contains at least two calibrated sequences around the metal implants. One sequence is a standard CT scan, typically showing severe metal artifacts that hinder doctors' interpretation of details. The other sequence, obtained using GSI technology, effectively reduces artifacts by exploiting multi-energy X-rays.
This new dataset enables us to train the models on real metal artifact data, thereby narrowing the gap between experimental results and real-world applications. 

Secondly, we introduce a novel method that efficiently reduces metal artifacts while preserving the original information. Our approach is based on an in-depth analysis of artifact impact. Despite metal implants absorbing significant radiation and obscuring surrounding areas, the original CT slices retain substantial tolerance, preserving critical information~\cite{watzke2004pragmatic, boas2011evaluation}. Moreover, the 3D reconstruction nature of CT imaging provides redundant information in adjacent spaces, which can be used to recover missing details. By leveraging these existing data sources, we can restore artifact-affected areas without introducing extraneous information.
To validate this observation, we manually labeled implants and important objects in the artifact areas. We then trained two segmentation models to predict these masks from both the artifact-affected ordinary sequences and the GSI sequences. The results demonstrate that the ordinary CT sequence can achieve similar accuracy [mIoU: 0.9213] to GSI data [mIoU: 0.9265], despite human difficulty in clearly delineating accurate boundaries in the ordinary sequences. This validation indicates that ordinary CT sequences contain sufficient information to match the detail provided by GSI data within artifact regions. The challenge lies in the different representations, which make ordinary CT slices harder to interpret.

Based on these observations, we propose the Latent GSI Alignment Framework (LGA) as shown in \cref{fig:motivation} (b) and (c). Rather than generating unclear information, our approach aligns the representations of ordinary CT sequences with those of GSI CT sequences. We achieve this by first converting both ordinary and GSI CT sequences into latent spaces using a Variational Autoencoder (VAE)~\cite{kingma2013auto}—an encoder model that compresses input data to a compressed latent space while capturing the most essential features. 
This makes the latent space ideal for aligning representations of ordinary and GSI CT slices. We then employ an alignment network that takes the latent codes of the ordinary CT as input and adjusts them to match the latent codes derived from corresponding GSI CT slices. During training, we use an Information Invariant Loss to ensure the alignment network preserves the original information. Finally, to accurately decode the latent codes into clean and readable CT slices, we use a VAE decoder to convert the latent codes back into CT space.
In summary, this paper presents four key contributions:

\begin{itemize}
    \item We present the Latent GSI Alignment Framework, a novel approach that effectively reduces metal artifacts in CT slices without introducing erroneous information. 
    \item We develop a novel Artifact-reducing VAE to achieve effective latent code encoding and decoding for the metal artifact reduction task. Compared with ordinary VAE, our method fully exploits the structural information in CT sequences and decouples the representation decoding for clear image representation.
    \item We introduce the ArtifactGSI dataset, which contains real data collected from patients who have received orthopedic surgery with artificial metal implants. This dataset bridges the domain gap between artifact reduction model training and real applications.
    \item We establish a new benchmark and assess metal artifact reduction methods.
    Experimental results demonstrate that our proposed method effectively reduces artifacts, outperforming state-of-the-art algorithms.
\end{itemize}

\begin{figure*}
    \centering
    \includegraphics[width= 0.98 \linewidth]{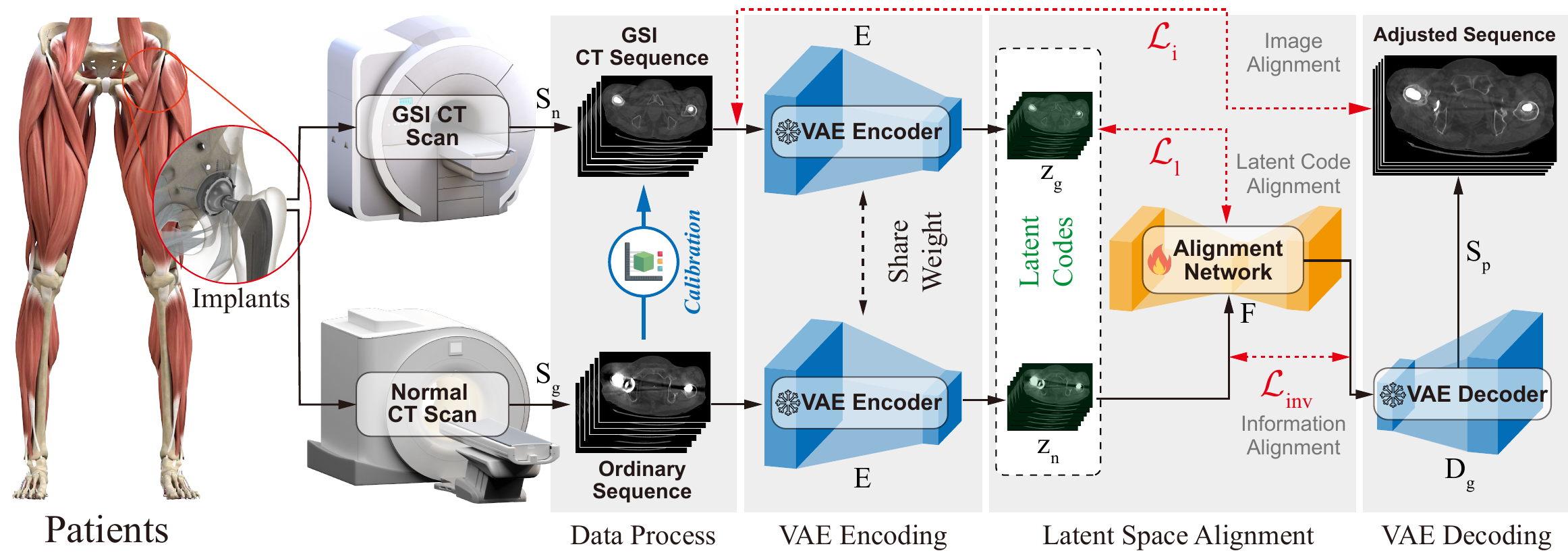}
    \caption{\textbf{Illustration of the Proposed Latent Space Alignment Framework.} The pipeline consists of four stages: Data Processing, VAE encoding, Latent Space Alignment, and VAE decoding.}
    \label{fig:framework}
    \vspace{-3mm}
\end{figure*}

\section{Related Works}
Metal artifact removal in CT imaging has been a long-standing challenge problem \cite{gjesteby2016metal}.
Traditional methods aim to directly correct the physical effects such as beam hardening~\cite{park2015metal, zhang2010beam} and photon starvation~\cite{kachelriess2001generalized} that cause artifacts. The methods usually fail to achieve satisfactory results because the signal received by X-ray detectors is severely disrupted. Some methods treat the metal-affected regions on the sinogram as missing areas and attempt to use techniques such as linear interpolation (LI)~\cite{kalender1987reduction} to restore these regions. The normalized metal artifacts reduction (NMAR)~\cite{meyer2010normalized} applies a tissue classification based on specific thresholds on the artifact-affected image or the LI-corrected image to remove artifacts and produce a prior image. This kind of restoration-based method always introduces new artifacts. Other methods remove metal artifacts by iteratively reconstructing the images from a series of projections~\cite{wang1996iterative, wang1999iterative}, these methods often require a large amount of computational resources and take a long time to produce results.

The deep learning-based metal artifacts removal methods~\cite{gjesteby2017deep, park2017sinogram, wu2023unsupervised, lee2024neural} can be categorized into three main types. The first category is sinogram domain-based methods, which adopt deep learning to restore the sinogram~\cite{liao2019generative, ghani2019fast}.
The second is image domain-based methods. ADN~\cite{liao2019adn} is the first network based on an unsupervised approach to disentangle metal artifacts in the latent space just using unpaired CT images. DICDNet~\cite{wang2021dicdnet} builds on the specific morphological characteristics of metal artifacts, treating them as inherent prior knowledge, to design an interpretable deep convolutional dictionary network. Similarly, the methods such as ACDNet~\cite{wang2022adaptive}, OSCNet~\cite{wang2022orientation} and OSCNet+~\cite{wang2023oscnet} are image domain-based. However, these methods are trained on synthetic datasets, which have a significant domain gap compared to real clinical CT images.
The third is dual domain-based methods. DuDoNet~\cite{lin2019dudonet} is the first end-to-end dual-domain network for metal artifacts reduction. InDuDoNet~\cite{wang2021indudonet} joints the image domain and sinogram domain to build an interpretable reconstruction model. Besides, MEPNet~\cite{wang2023mepnet}, InDuDoNet+\cite{wang2023indudonet+} and DuDoDp~\cite{liu2024unsupervised} are also dual domain-based methods. 

\section{Method}
In this section, we provide a comprehensive introduction to our method. First, in Sec.~\ref{Sec:framework}, we propose the overall pipeline of our Latent Space Alignment Framework. In Sec.~\ref{Sec:vae}, we introduce the newly designed Artifact-Reducing VAE structure. Next, in Sec.~\ref{Sec:network}, we detail the alignment network. Finally, in Sec.~\ref{Sec:loss}, we discuss the loss function used in this framework.

\subsection{Latent Space Alignment Framework}
\label{Sec:framework}
As shown in \cref{fig:framework}, our pipeline begins with two CT sequences captured from the same patient: an ordinary CT sequence $S_n$ and a GSI CT sequence $S_g$.  Notably, we capture the two sequences simultaneously using a GE Discovery CT 750HD device, which is equipped with multiple imaging systems capable of producing multiple types of CT sequences at the same time. This approach ensures that the patient’s condition remains identical in both sequences, eliminating any movement-related discrepancies.
We first apply a calibration algorithm to align the two sequences. Specifically, we employ a similarity calculation algorithm to match slices between the two sequences, identifying pairs that represent the same positions:
\begin{equation}
 \text{Match}({S}_n[i], {S}_g[j]) = \max_{i,j} \left[ \sum_{} (S_n[i] \ast S_g[j]) \right],
\end{equation}
where ${S}_g[j]$ is the corresponding slice in the GSI sequence for ${S}_n[i]$.
Next, we use a scale-adjusting algorithm to adjust the ordinary CT slices so that can match pixel-wise with the GIS CT slices:
\begin{equation}
\hat{S}_n[i] = \text{AffinityTransform}({S}_n[i], {S}_g[j]).
\end{equation}
%
Following this procedure, $\hat{S}_n[i]$ and ${S}_g[j]$ represent identical regions of the patient at the same scale, differing only in their representation. To adjust the representation of $\hat{S}_n[i]$ to match the clear representation of ${S}_g[j]$, we then transform these sequences into latent codes using the VAE encoder $E$:
\begin{equation}
z_n[i] = E(\hat{S}_n[i]), \quad z_g[j] = E(S_g[j]).
\end{equation}
where $z_n[i]$ and $z_g[j]$ represent the latent codes for the ordinary and GSI sequences, respectively. We employ the alignment network $F$ to adjust these latent codes. This network takes the latent code of the ordinary CT, $z_n$, as input and produces adjusted ones $\bar{z}_n$ that are expected to closely resemble the $z_g$. 
This adjustment can be expressed as:
\begin{equation}
\bar{z}_n = F(z_n).
\end{equation}

Finally, we employ a VAE Decoder $D_g$ to convert the adjusted latent codes back into the CT sequence space. This decoder, trained on GSI and clean CT sequences, generates clear, easily interpretable CT images. We will delve into the details of this process in Sec.~\ref{Sec:vae}. This conversion is expressed by the equation:
\begin{equation}
S_p[i] = D_g(\bar{z}_n[i]).
\end{equation}
In this equation, $S_p$ represents the reconstructed CT sequences with minimized artifact representations and improved structural details. These improvements make the sequences suitable for diagnosis like GSI data.

\subsection{Artifact-reducing VAE}
\label{Sec:vae}
\begin{figure}[t]
    \centering
\includegraphics[width=0.48\textwidth]{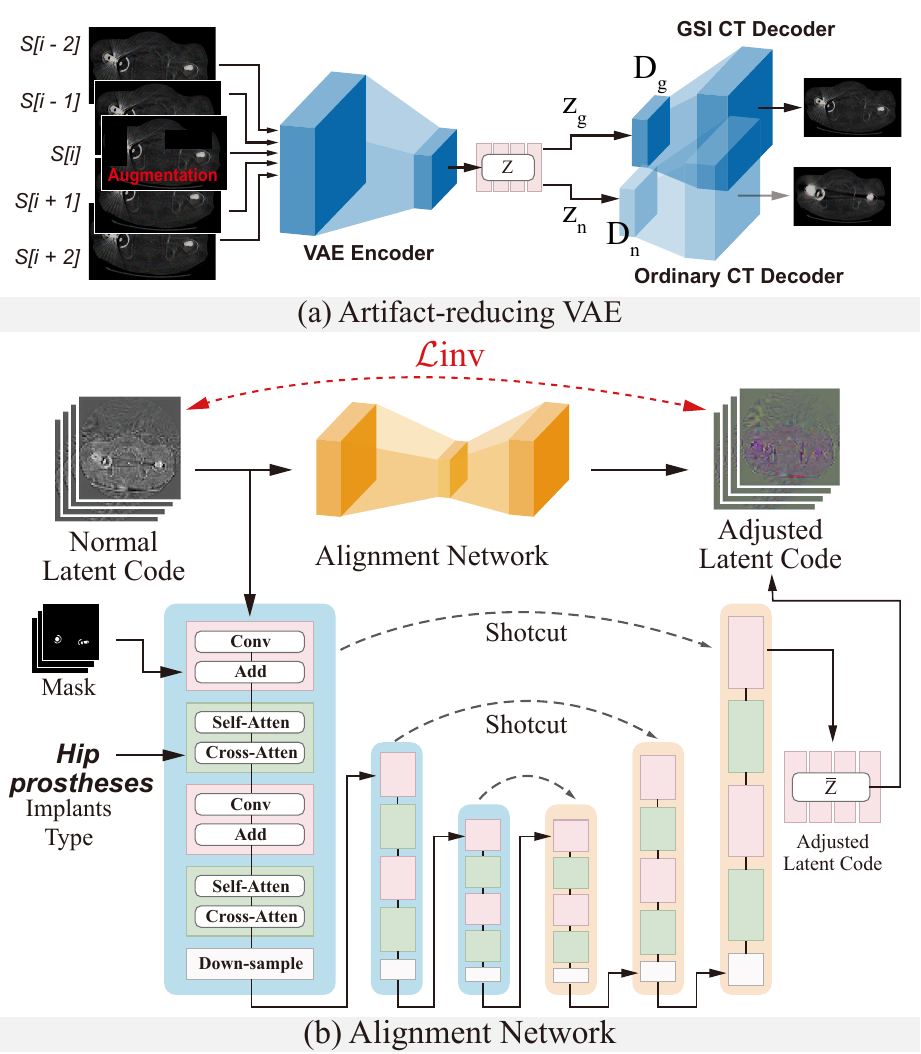} 
    \caption{\textbf{Illustration of the Proposed Artifact-reducing VAE and Alignment Network}. \textbf{(a)} Artifact-reducing VAE: The VAE Encoder takes CT volumes as input, using data augmentation on the center slice to enhance information aggregation. Two VAE decoders are employed: one trained on ordinary data and another on clean data, decoupling the decoding representations for artifact-affected and clean images. \textbf{(b)} Alignment Network: The network employs an encoder-decoder structure, enhanced with additional input signals and transformer-residual modules.}
    \label{fig:vae_network}
     \vspace{-3mm}
\end{figure}

%
By incorporating several significant improvements, we propose a novel VAE structure called Artifact-reducing VAE.
The first improvement focuses on the encoder design. As illustrated in \cref{fig:vae_network} (a), our encoder differs from a standard VAE encoder structure. 
While a typical encoder processes a single image to generate corresponding latent codes, our approach takes a 3D volume as input comprising several adjacent slices from a CT sequence:
\begin{equation}
V[i] = \{ S[i-k], \ldots, S[i], \ldots, S[i+k] \},
\end{equation}
where $V[i]$ represents a volume centered on slice $S[i]$, encompassing up to $k$ adjacent slices on each side. 
Then the encoder then generates latent codes solely for this central slice within the volume:
\begin{equation}
z[i] = E(V[i]).
\end{equation}

This design can leverage complementary information from adjacent slices to compensate for artifact-affected areas, aiding in the recovery of original structures. To improve the effectiveness of volumetric input, we implement a \textit{volume aggregation data augmentation strategy}. 
During training, we apply various augmentations—such as random masking, blurring, and other information down-sampling techniques—to the central slice $S[i] \in V[i]$. However, the reconstruction loss is still computed using the original unaltered $S[i]$. This approach compels the encoder to learn complementary information from surrounding slices, thereby improving its ability to accurately reconstruct artifact-influenced areas.

The second improvement involves designing an asymmetric architecture, as shown in \cref{fig:vae_network} (a). Our structure utilizes a common encoder but employs two distinct decoders for different types of CT sequences.
During the VAE training phase, both the ordinary CT ($\hat{S}_n$) and the GSI CT ($S_g$) are processed through the same encoder $E$ to generate latent codes $z_n$ and $z_g$. Subsequently, two separate decoders—$D_n$ for ordinary CT and $D_g$ for GSI CT—decode these latent codes back into the image space.
As mentioned in the previous section, we only employ the decoder $D_g$ during the artifact-reduction process. This structure effectively decouples the image space of artifact-affected ordinary CT slices from that of clean images. Since $D_g$ is trained on highly readable data, it can minimize the impact of artifacts during decoding.

To further eliminate subtle artifacts in GSI slices, we propose an additional training strategy to enhance the decoder's performance. As shown in \cref{fig:framework}, while GSI technology significantly reduces artifacts in CT slices, subtle imperfections can persist, potentially obscuring critical details for doctors' interpretation. 
To improve readability and produce even higher-quality images than the original GSI scans, we adjust the training data for the VAE decoders. When training $D_g$, we use a small portion (25\%) of GSI slices, with the remaining 75\% consisting of clean CT slices from patients without implants, providing completely artifact-free images. 
During training, the VAE learns to select essential information, focusing on features common to both clean and GSI slices while naturally ignoring trivial artifacts in GSI slices.
Consequently, when we use $D_g$ to decode the aligned hidden code $\bar{z}_n$, it produces images of superior quality compared to the original GSI scans, offering improved clarity and readability.

\subsection{Alignment Network}
\label{Sec:network}
\cref{fig:vae_network} (b) illustrates the structure of the alignment network. We employ a U-Net~\cite{ronneberger2015u} architecture as our base model. To enhance alignment performance, we introduce two key improvements. First, we integrate Residual Blocks and Transformer Blocks within both the U-Net Encoder and Decoder. These block types offer complementary advantages for the alignment process. Residual Blocks excel at processing local spatial information, helping consolidate details in artifact-affected regions. Transformer Blocks, on the other hand, gather information from a global perspective, facilitating the transformation of feature representations from ordinary CT to those resembling GSI.

Secondly, we enhance the alignment network with two types of prior information. The first is the implant type, $I$, which can be easily obtained from patients' medical records. When unavailable, it is marked as ``None". We convert this implant type into one-hot embeddings, then fuse them with visual features in the Transformer Block, boosting the network's context awareness. The second prior is the metal mask within the CT slice. We compute this by filtering regions with high Hounsfield Unit (HU) values in CT slices. Metal implants, denser than other human organs, yield higher HU values. By identifying these high-HU areas, we pinpoint metal implant positions, providing valuable prior information:
\begin{equation}
M(x, y) = 
\begin{cases} 
1 & \text{if } \text{HU}(x, y) > \text{threshold} \\
0 & \text{otherwise}.
\end{cases}
\end{equation}
where $M(x, y)$ indicates whether a pixel at position $(x, y)$ is part of the metal, based on a predefined HU threshold. This mask serves as prior information in the alignment network, highlighting regions that produce metal artifacts. 

To effectively exploit these priors, we merge the metal mask with the visual features in the Residual Blocks. The entire operation of the alignment network $F$ can be summarized by the following formula:
\begin{equation}
\begin{aligned}
    \bar{z}_n &= F(z_n)= \text{ALU-Net} (z_n, I, M).
\end{aligned}
\end{equation}


\subsection{Information Invariant Loss}
\label{Sec:loss}
A crucial aspect of the metal artifact reduction method is preserving structural information while avoiding the introduction of erroneous data into CT sequences. Our approach, which does not rely on random noise-based generation algorithms, inherently minimizes the risk of introducing informational noise. To further ensure information consistent during the artifact reduction process, we incorporate an Information Invariant Loss $L_{inv}$ into our pipeline.
To accomplish this, we first employ two encoders, $Ec_n$ and $Ec_g$, to transform the latent codes from the ordinary CT sequences $z_n$ and the GSI CT sequences $z_g$ into high-level feature maps $f_n$ and $f_g$, respectively:
\begin{equation}
f_n = Ec_n(z_n), \quad f_g = Ec_g(z_g).
\end{equation}

We then employ a contrastive learning approach to train the encoders. Specifically, we pool the feature maps of each slice into one-dimensional embeddings and then minimize the distance between embeddings from ordinary CT slices and GSI slices of the same position. Simultaneously, we maximize the distance between embeddings from different patients or different positions within the same patient:
\begin{equation}
\begin{aligned}
D_{\text{min}} &= \sum_i \| \text{pool}(f_n[i]) - \text{pool}(f_g[i]) \|^2, \\
D_{\text{max}} &= \sum_i \sum_{j \neq i} \| \text{pool}(f_n[i]) - \text{pool}(f_g[j]) \|^2, \\
(Ec_n^*, Ec_g^*) &= \arg\min_{Ec_n^*, Ec_g^*} \left( D_{\text{min}} - \gamma D_{\text{max}} \right).
\end{aligned}
\end{equation}
where $Ec_n^*$ and $Ec_g^*$ denote the parameters of the encoders and $\text{pool}$ represents the pooling operation that reduces feature maps to one-dimensional embeddings, and $\gamma$ is a weighting factor to balance the loss terms. During the training phase of the alignment network, we keep the encoders $Ec_{n}$ and $Ec_{g}$ fixed and use them to ensure information invariance during the alignment process.

Specifically, we extract features from the original input latent codes and the latent codes produced by the alignment network. We then minimize the difference between these two feature maps in a pixel-wise manner:
\begin{equation}
\begin{aligned}
L_{\text{inv}} = \sum_i \| Ec_{n}(z_n[i]) - Ec_{g}(\bar{z}_n[i]) \|^2
\end{aligned}
\end{equation}
%

This loss ensures that the alignment network modifies only the representation of the CT slices without introducing any extraneous information. 
Finally, the overall loss function used to train the alignment network can be formulated as:
\begin{equation}
\begin{aligned}
L = L_{i} + \alpha L_{l} + \beta L_{inv}
\end{aligned}
\end{equation}
where $L_{i}$ is the MSE loss between the target and prediction in the image space and $L_{l}$ is the mse loss in the latent space.

\section{The Proposed ArtifactGSI Dataset}
To demonstrate the effectiveness of the proposed Metal Artifact Reducing Framework in medical applications, we introduce a new dataset named the ArtifactGSI dataset. This dataset comprises data from 157 different patients. Among them, 115 patients have undergone at least one implant replacement surgery and currently have metal implants in their bodies. The remaining 42 patients do not have any metal implants and are primarily used to train the VAE decoder mentioned in Sec.~\ref{Sec:vae}.

For patients with implants, we collect at least two CT sequences per patient: one from an ordinary CT scan and another from a GSI CT scan. The sequences cover six body regions: thoracic and lumbar vertebrae, pelvis, hip joint, femur, and knee joint. The dataset includes three main types of implants: hip prostheses, fracture internal fixations, and spinal internal fixations. Furthermore, it encompasses a variety of patient conditions. The examples from the dataset and the detailed patient distribution are provided in the supplementary material.


\noindent \textbf{Benchmark.} To comprehensively evaluate the performance of the proposed framework, we introduce a new benchmark for the CT Metal Artifact Reduction task. Our evaluation is defined across three key aspects.
1) \textbf{Effectiveness in Sub-tasks:} Since metal artifacts reduction methods must enhance diagnostic capabilities, we include a common subtask—semantic segmentation. We employ several senior doctors to manually label the masks for five classes on the GSI CT images. We then assess the performance of semantic segmentation models, trained on GSI data, using the artifact-reduced normal CT inputs. This metric helps indicate the quality of artifact reduction while evaluating if there is noise information introduced.
2) \textbf{Similarity to GSI Frames:} We directly assess the similarity between GSI frames and artifact-reduced normal CT frames by calculating metrics such as Peak Signal-to-Noise Ratio (PSNR), Structural Similarity Index (SSIM)~\cite{wang2004image} and Learned Perceptual Image Patch Similarity (LPIPS)~\cite{zhang2018unreasonable}. However, it is worth noting that our method can further eliminate detailed artifacts and potentially provide even better quality than the GSI data, so these metrics should be considered as a partial reference.
3) \textbf{Expert Study:} We enlist several experienced doctors to score the artifact-reduced CT frames. Since these processed images are often interpreted by doctors in real applications, doctor scores are the most critical metric. The doctors provide scores in two aspects: correctness and readability. Correctness assesses whether the information in the CT frames accurately reflects the patient's condition. Readability measures the subjective ease with which doctors can interpret the images.
4) \textbf{Generalization:} To assess the generalization of our method across different device types, we created an additional generalization test set. This set includes patient data with real artifacts collected from common ordinary CT devices by manufacturers such as Siemens, Philips, United Imaging and data from the open-source SpineWeb dataset~\cite{glocker2013vertebrae}.  These devices, unlike the training set devices (GE), come from different manufacturers and have not undergone any registration procedures. This highlights the practical generalization capability of our method.
Notably, because these data were acquired from ordinary CT devices, there is no ground truth available for quantitative evaluation, so we performed a qualitative comparison instead.

\section{Experiments}


\subsection{Implementation Details}
\label{Sec:imp_detail}
We conduct experiments on four NVIDIA RTX A6000 GPUs, and implement the framework with PyTorch~\cite{paszke2019pytorch} and diffusers library~\cite{von-platen-etal-2022-diffusers}. We adopt the AdamW~\cite{loshchilov2017decoupled} optimizer and set the learning rate to \(1 \times 10^{-5}\), and train for 10 epochs, which takes about 30 hours. For the Alignment Network, we use the same optimizer and learning rate configuration as the Artifact-reducing VAE, training for 50 epochs, which takes about 60 hours. For hyperparameter settings of the loss functions, we set \(\alpha_1 = 1, \beta = 0.001\).


For the dataset split, we use 80\% patients for training, with the remaining patients used for testing. In the comparative experiments, we use the experimental parameter settings from the original method, including the input and output resolution and follow the simulation protocol in~\cite{yu2020deep, wang2021indudonet} to generate the sinogram corresponding to each CT image. When calculating the final metrics, we uniformly rescale the images to a size of \(512 \times 512\) to evaluate the effectiveness of artifact reduction.

\begin{figure*}[ht]
    \centering
    \includegraphics[width= 1.0 \textwidth]{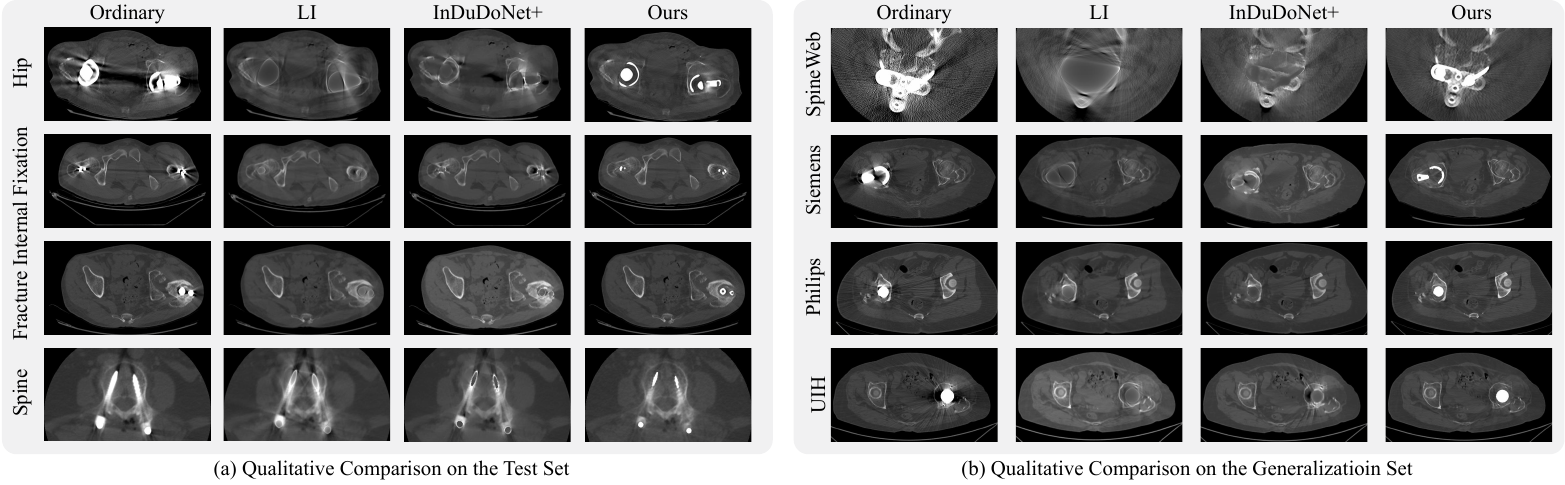}
    \caption{\textbf{Qualitative Comparisons.} (a) Comparison on the Test Set: Images of patients with hip prostheses used in total hip arthroplasty, fracture internal fixation, and spinal internal fixation. (b) Comparison on the Generalization Set: Evaluation on data from unseen CT machines (SpineWeb dataset, Siemens, Philips, and UIH CT machines) to demonstrate generalization.}
    \label{fig:visualize}
    \vspace{-4mm}
\end{figure*}

\subsection{Ablation Study}
\label{Sec:ablation}
In Table \ref{tab:ablation_study}, we conduct ablation studies on the proposed Latent Space Alignment Framework. First, we use a 2D VAE with a single decoder, and an Alignment Network without additional input signals or transformer-residual modules as baseline. Next, we introduce the proposed Artifact-reducing VAE (ARVAE), but we still use the incomplete Alignment Network mentioned above for latent space alignment. Here, we observe improvements in two metrics, with PSNR increasing by 0.3191, and SSIM increasing by 0.0098. Then, we incorporate the Information Invariant Loss, which results in slight performance gains. Additionally, we add the complete Alignment Network (AN) to the Artifact-reducing VAE separately, without the Information Invariant Loss, and observe a significant performance improvement, with PSNR increasing by 3.8965 and SSIM increasing by 0.0119,  indicating that additional input signals and transformer-residual modules play an important role in aligning the latent codes. Finally, the full version, achieve optimal performance, with PSNR reaching 37.9146, SSIM reaching 0.9690 and LPIPS reaching 0.0198.

\begin{table}[t]
    \centering
    \caption{\textbf{Ablation Study}. The results using different components with the Latent Space Alignment Framework. \textbf{ARVAE} represents the Artifact-reducing VAE. \textbf{AN} represents the Alignment Network. \textbf{Loss} represents the Information Invariant Loss.}
    \setlength{\tabcolsep}{10pt}
\resizebox{\linewidth}{!}{
    \begin{tabular}{c|c|c|c|c|c}
        \toprule
        \multicolumn{3}{c|}{\textbf{Components}} & \multirow{2}{*}{\textbf{PSNR $\uparrow$}} & \multirow{2}{*}{\textbf{SSIM $\uparrow$}} & \multirow{2}{*}{\textbf{LPIPS $\downarrow$}} \\
        \cline{1-3} 
        \textbf{ARVAE} & \textbf{AN} & \textbf{Loss} & ~ & ~ \\
        \hline
        \hline
        ~ & ~ & ~ & 32.8084 & 0.9470 & 0.0326 \\
        \checkmark & ~ & ~ & 33.1275 & 0.9568 & 0.0346 \\
        \checkmark & ~ & \checkmark & 33.3838 & 0.9579 & 0.0326 \\
        \checkmark & \checkmark & ~ & 37.0240 & 0.9687 & 0.0215 \\
        \checkmark & \checkmark & \checkmark & \textbf{37.9146} & \textbf{0.9690} & \textbf{0.0198} \\
        \bottomrule
    \end{tabular}}
    \label{tab:ablation_study}
\end{table}

\subsection{Semantic Segmentation}
\label{Sec:segmentation}
To assess whether the removal of artifacts affects the inherent anatomical structures in the CT images, we incorporated semantic segmentation as a downstream task.
In this task, orthopedic experts manually annotate segmentation masks for the six most representative regions in the ArtifactGSI dataset: metal implants, left and right femur, and left and right pelvis. We utilize the GSI sequence for training, which effectively suppresses metal artifacts while preserving consistent and clear anatomical structures. Then, we conduct testing on our method and other methods to evaluate the segmentation performance of different approaches on these key anatomical structures, thereby reflecting whether these results can maintain the same readability and correctness as the GSI sequence.

From Table~\ref{tab:semantic_segmentation}, it can be seen that other methods perform poorly in segmenting metallic regions in CT images, as they fail to restore these areas. While removing artifacts, these methods also introduce adverse effects on the clear representation of surrounding tissues. As a result, for tissues slightly distant from the metal implants, although segmentation metrics have improved, there is still a noticeable gap compared to the GSI sequence. In contrast, our method not only effectively restores the metal implant region but also maximally preserves the clarity of the surrounding structures, with the segmentation metric of mean Intersection over Union (mIoU) differing by no more than 0.05 compared to the GSI sequence.

\begin{table}[t]
    \centering
    \caption{\textbf{Semantic Segmentation Performance}. Mean Intersection over Union (mIoU) values of the segmentation model for six key anatomical structures across different methods.}
    \setlength{\tabcolsep}{8pt}
    \resizebox{\linewidth}{!}{
    \begin{tabular}{c|c|c|c|c|c}
        \toprule
         \makecell{\textbf{Body Part} \\ \textbf{(mIoU)}} & \makecell{\textbf{Metal} \\ \textbf{Implants}} & \makecell{\textbf{Left} \\ \textbf{Pelvis}} & \makecell{\textbf{Right} \\ \textbf{Pelvis}} & \makecell{\textbf{Left} \\ \textbf{Femur}} & \makecell{\textbf{Right} \\ \textbf{Femur}} \\
        \hline
        \rowcolor{gray!10} GSI & 0.8944 & 0.9273 & 0.9115 & 0.9442 & 0.9302 \\ 
        \hline
        \hline
        LI\cite{kalender1987reduction} & 0.0998 & 0.7652 & 0.6513 & 0.6863 & 0.5584  \\
        \hline
        DuDoDp\cite{liu2024unsupervised} & 0.0491 & 0.7607 & 0.7023 & 0.6027 & 0.5523 \\
        \hline
        OSCNet\cite{wang2022orientation} & 0.1059 & 0.7892 & 0.6612 & 0.7336 & 0.8098 \\ 
        \hline
        OSCNet+\cite{wang2023oscnet} & 0.0779 & 0.7842 & 0.6289 & 0.6902 & 0.7965 \\ 
        \hline
        Ours & \textbf{0.8582} & \textbf{0.8999} & \textbf{0.9059} & \textbf{0.9232} & \textbf{0.9013} \\ 
        \bottomrule
    \end{tabular}}
    \label{tab:semantic_segmentation}
    \vspace{-4mm}
\end{table}

\subsection{Image Quality}
\label{Sec:image_quality}
Table~\ref{tab:comparison_1} shows quantitative evaluations of the image quality after artifact removal using different methods. We select traditional method linear interpolation (LI)~\cite{kalender1987reduction}, dual domain-based method DuDoDp~\cite{liu2024unsupervised}, image domain-based method OSCNet~\cite{wang2022orientation} and OSCNet+~\cite{wang2023oscnet} for comparison. Our method significantly outperforms these approaches in all three metrics. Notably, to demonstrate the effectiveness of our real-domain training data, the other methods were evaluated using their released pretrained parameters without fine-tuning on our new dataset. Moreover, to highlight the intrinsic improvements of our method, we also include a comparison in the supplementary materials where all approaches are fine-tuned on our dataset. In this scenario, our method still significantly outperform the SOTA methods.

\cref{fig:visualize} (a) illustrates three types of metal implants. In CT slices with large, bilateral metal regions, the standard sequence exhibits a distinct dark band between the implants, resulting in missing information. Similarly, screw-like implants, although smaller, generate star-shaped artifacts that distort the surrounding tissue. Traditional methods and those trained on synthetic datasets can partially remove these artifacts and restore affected areas, but they often fail to accurately reconstruct the metal implant regions, leading to diminished visual quality. In contrast, our method effectively eliminates these artifacts while preserving the structural details of the implants and restoring the compromised regions. For visualization, we revert the outputs of all models to their original data range using appropriate normalization techniques, and display them with a window level of 500 and a window width of 2000.
To demonstrate the generalizability of our approach, we also conducted qualitative comparisons using various ordinary CT machines in \cref{fig:visualize} (b). As shown in the figure, even when applied to CT devices not seen during training, our method maintains desired performance, effectively reducing artifacts while preserving structural information. Notably, since these data were acquired from normal CT devices, no ground truth is available for quantitative comparison.

\begin{table}[t]
    \centering
    \caption{\textbf{Quantitative Evaluation}. Comparison with different methods on the ArtifactGSI dataset.}
    \setlength{\tabcolsep}{15pt}
\resizebox{\linewidth}{!}{
    \begin{tabular}{c|c|c|c}
        \toprule
        \textbf{Methods} & \textbf{PSNR $\uparrow$} & \textbf{SSIM $\uparrow$} & \textbf{LPIPS $\downarrow$} \\ 
        \hline
        \hline
        LI~\cite{kalender1987reduction} & 32.5903 & 0.9638 & 0.0457 \\
        \hline
        DuDoDp~\cite{liu2024unsupervised} & 31.2263 & 0.9569 & 0.0510 \\
        \hline
        OSCNet~\cite{wang2022orientation} & 31.4065 & 0.9612 & 0.0439 \\
        \hline
        OSCNet+~\cite{wang2023oscnet} & 31.3298 & 0.9632 & 0.0426 \\
        \hline
        Ours & \textbf{37.9146} & \textbf{0.9690} & \textbf{0.0198} \\
        \bottomrule
    \end{tabular}}
    \label{tab:comparison_1}
    \vspace{-4mm}
\end{table}

\subsection{Expert Study}
\label{Sec:human_evaluation}
%
We organize an expert study to assess the quality of artifact removal in the images from the doctor's perspective. We distribute the results from the test set to several medical experts, ensuring that each image is evaluated by at least two doctors. The experts are blind to the source of the images and independently rate on a scale of 1 to 5, according to the criteria mentioned in \cite{zhang2021reduction}. We average the ratings, and the results are shown in Table~\ref{tab:human_evaluation}. 

\begin{table}[t]
    \centering
    \caption{\textbf{Human Evaluation Results}. Medical experts rate the results obtained from different methods based on clinical standards.}
    \setlength{\tabcolsep}{20pt}
    \resizebox{\linewidth}{!}{
    \begin{tabular}{c|c|c}
        \toprule
        \textbf{Methods} & \textbf{Readability} & \textbf{Correctness} \\
        \hline
        \hline
        LI~\cite{kalender1987reduction} & 1.58 & 2.87 \\
        \hline
        DuDoDp~\cite{liu2024unsupervised} & 1.55 & 2.59 \\
        \hline
        OSCNet~\cite{wang2022orientation} & 2.21 & 3.58 \\
        \hline
        OSCNet+~\cite{wang2023oscnet} & 2.15 & 3.90 \\
        \hline
        Ours & \textbf{4.47} & \textbf{4.23} \\ 
        \bottomrule 
    \end{tabular}
    }
    \label{tab:human_evaluation}
    \vspace{-4mm}
\end{table}

\section{Conclusion}
In this paper, we introduce the Latent GSI Alignment Framework to address the reduction of metal artifacts in CT sequences. Our work begins with a detailed analysis of artifact-affected areas, revealing that, despite being unreadable, these areas contain sufficient information to reconstruct structures. However, these sequences lack effective representation to clearly depict these regions. To resolve this, we develop a framework that aligns ordinary CT slices with GSI CT slices in latent space.
Our approach includes the design of a novel VAE structure that fully utilizes the spatial information encoded in adjacent 3D volumes, decoding latent codes into clearer image spaces. We also propose an alignment network that leverages prior information about metal area masks and implant types. Additionally, we introduce an information invariant loss to prevent the introduction of extraneous noise.
To validate our approach, we create a new dataset and establish a comprehensive benchmark, comprising both standard CT and GSI CT sequences. Experimental results demonstrate that our method effectively reduces metal artifacts while preserving the integrity of the input data.

{
    \small
    \bibliographystyle{ieeenat_fullname}
    \bibliography{main}

\begin{thebibliography}{54}
\providecommand{\natexlab}[1]{#1}
\providecommand{\url}[1]{\texttt{#1}}
\expandafter\ifx\csname urlstyle\endcsname\relax
  \providecommand{\doi}[1]{doi: #1}\else
  \providecommand{\doi}{doi: \begingroup \urlstyle{rm}\Url}\fi

\bibitem[Boas and Fleischmann(2011)]{boas2011evaluation}
F~Edward Boas and Dominik Fleischmann.
\newblock Evaluation of two iterative techniques for reducing metal artifacts in computed tomography.
\newblock \emph{Radiology}, 259\penalty0 (3):\penalty0 894--902, 2011.

\bibitem[Boas et~al.(2012)Boas, Fleischmann, et~al.]{boas2012ct}
F~Edward Boas, Dominik Fleischmann, et~al.
\newblock Ct artifacts: causes and reduction techniques.
\newblock \emph{Imaging Med}, 4\penalty0 (2):\penalty0 229--240, 2012.

\bibitem[De~Man et~al.(1998)De~Man, Nuyts, Dupont, Marchal, and Suetens]{de1998metal}
Bruno De~Man, Johan Nuyts, Patrick Dupont, Guy Marchal, and Paul Suetens.
\newblock Metal streak artifacts in x-ray computed tomography: a simulation study.
\newblock In \emph{1998 IEEE Nuclear Science Symposium and Medical Imaging Conference}, pages 1860--1865. IEEE, 1998.

\bibitem[Ghani and Karl(2019)]{ghani2019fast}
Muhammad~Usman Ghani and W~Clem Karl.
\newblock Fast enhanced ct metal artifact reduction using data domain deep learning.
\newblock \emph{IEEE Transactions on Computational Imaging}, 6:\penalty0 181--193, 2019.

\bibitem[Gjesteby et~al.(2016)Gjesteby, De~Man, Jin, Paganetti, Verburg, Giantsoudi, and Wang]{gjesteby2016metal}
Lars Gjesteby, Bruno De~Man, Yannan Jin, Harald Paganetti, Joost Verburg, Drosoula Giantsoudi, and Ge Wang.
\newblock Metal artifact reduction in ct: where are we after four decades?
\newblock \emph{IEEE Access}, 4:\penalty0 5826--5849, 2016.

\bibitem[Gjesteby et~al.(2017)Gjesteby, Yang, Xi, Zhou, Zhang, and Wang]{gjesteby2017deep}
Lars Gjesteby, Qingsong Yang, Yan Xi, Ye Zhou, Junping Zhang, and Ge Wang.
\newblock Deep learning methods to guide ct image reconstruction and reduce metal artifacts.
\newblock In \emph{Medical Imaging 2017: Physics of Medical Imaging}, pages 752--758. SPIE, 2017.

\bibitem[Glocker et~al.(2013)Glocker, Zikic, Konukoglu, Haynor, and Criminisi]{glocker2013vertebrae}
Ben Glocker, Darko Zikic, Ender Konukoglu, David~R Haynor, and Antonio Criminisi.
\newblock Vertebrae localization in pathological spine ct via dense classification from sparse annotations.
\newblock In \emph{Medical Image Computing and Computer-Assisted Intervention--MICCAI 2013: 16th International Conference, Nagoya, Japan, September 22-26, 2013, Proceedings, Part II 16}, pages 262--270. Springer, 2013.

\bibitem[Hounsfield(1973)]{hounsfield1973computerized}
Godfrey~N Hounsfield.
\newblock Computerized transverse axial scanning (tomography): Part 1. description of system.
\newblock \emph{The British Journal of Radiology}, 46\penalty0 (552):\penalty0 1016--1022, 1973.

\bibitem[Huang et~al.(2018)Huang, Wang, Tang, Zhong, and Zhang]{huang2018metal}
Xia Huang, Jian Wang, Fan Tang, Tao Zhong, and Yu Zhang.
\newblock Metal artifact reduction on cervical ct images by deep residual learning.
\newblock \emph{Biomedical Engineering Online}, 17:\penalty0 1--15, 2018.

\bibitem[Johnson et~al.(2007)Johnson, Krauss, Sedlmair, Grasruck, Bruder, Morhard, Fink, Weckbach, Lenhard, Schmidt, et~al.]{johnson2007material}
Thorsten~RC Johnson, Bernhard Krauss, Martin Sedlmair, Michael Grasruck, Herbert Bruder, Dominik Morhard, Christian Fink, Sabine Weckbach, Miriam Lenhard, Bernhard Schmidt, et~al.
\newblock Material differentiation by dual energy ct: initial experience.
\newblock \emph{European Radiology}, 17:\penalty0 1510--1517, 2007.

\bibitem[Kachelriess et~al.(2001)Kachelriess, Watzke, and Kalender]{kachelriess2001generalized}
Marc Kachelriess, Oliver Watzke, and Willi~A Kalender.
\newblock Generalized multi-dimensional adaptive filtering for conventional and spiral single-slice, multi-slice, and cone-beam ct.
\newblock \emph{Medical Physics}, 28\penalty0 (4):\penalty0 475--490, 2001.

\bibitem[Kalender et~al.(1987)Kalender, Hebel, and Ebersberger]{kalender1987reduction}
Willi~A Kalender, Robert Hebel, and Johannes Ebersberger.
\newblock Reduction of ct artifacts caused by metallic implants.
\newblock \emph{Radiology}, 164\penalty0 (2):\penalty0 576--577, 1987.

\bibitem[Kingma(2013)]{kingma2013auto}
Diederik~P Kingma.
\newblock Auto-encoding variational bayes.
\newblock \emph{arXiv preprint arXiv:1312.6114}, 2013.

\bibitem[Lee et~al.(2024)Lee, Ahn, and Baek]{lee2024neural}
Jooho Lee, Junhyun Ahn, and Jongduk Baek.
\newblock Neural attenuation fields for metal artifact reduction in dental ct.
\newblock In \emph{Medical Imaging 2024: Physics of Medical Imaging}, pages 173--179. SPIE, 2024.

\bibitem[Lee et~al.(2012)Lee, Park, Song, Kim, and Suh]{lee2012metal}
Young~Han Lee, Kwan~Kyu Park, Ho-Taek Song, Sungjun Kim, and Jin-Suck Suh.
\newblock Metal artefact reduction in gemstone spectral imaging dual-energy ct with and without metal artefact reduction software.
\newblock \emph{European Radiology}, 22:\penalty0 1331--1340, 2012.

\bibitem[Liao et~al.(2019{\natexlab{a}})Liao, Lin, Huo, Vogelsang, Sehnert, Zhou, and Luo]{liao2019generative}
Haofu Liao, Wei-An Lin, Zhimin Huo, Levon Vogelsang, William~J Sehnert, S~Kevin Zhou, and Jiebo Luo.
\newblock Generative mask pyramid network for ct/cbct metal artifact reduction with joint projection-sinogram correction.
\newblock In \emph{International Conference on Medical Image Computing and Computer-Assisted Intervention}, pages 77--85. Springer, 2019{\natexlab{a}}.

\bibitem[Liao et~al.(2019{\natexlab{b}})Liao, Lin, Zhou, and Luo]{liao2019adn}
Haofu Liao, Wei-An Lin, S~Kevin Zhou, and Jiebo Luo.
\newblock Adn: artifact disentanglement network for unsupervised metal artifact reduction.
\newblock \emph{IEEE Transactions on Medical Imaging}, 39\penalty0 (3):\penalty0 634--643, 2019{\natexlab{b}}.

\bibitem[Lin et~al.(2019)Lin, Liao, Peng, Sun, Zhang, Luo, Chellappa, and Zhou]{lin2019dudonet}
Wei-An Lin, Haofu Liao, Cheng Peng, Xiaohang Sun, Jingdan Zhang, Jiebo Luo, Rama Chellappa, and Shaohua~Kevin Zhou.
\newblock Dudonet: Dual domain network for ct metal artifact reduction.
\newblock In \emph{Proceedings of the IEEE/CVF Conference on Computer Vision and Pattern Recognition}, pages 10512--10521, 2019.

\bibitem[Liu et~al.(2024)Liu, Xie, Diao, Tan, and Liang]{liu2024unsupervised}
Xuan Liu, Yaoqin Xie, Songhui Diao, Shan Tan, and Xiaokun Liang.
\newblock Unsupervised ct metal artifact reduction by plugging diffusion priors in dual domains.
\newblock \emph{IEEE Transactions on Medical Imaging}, 2024.

\bibitem[Loshchilov(2017)]{loshchilov2017decoupled}
I Loshchilov.
\newblock Decoupled weight decay regularization.
\newblock \emph{arXiv preprint arXiv:1711.05101}, 2017.

\bibitem[Lyu et~al.(2020)Lyu, Lin, Liao, Lu, and Zhou]{lyu2020encoding}
Yuanyuan Lyu, Wei-An Lin, Haofu Liao, Jingjing Lu, and S~Kevin Zhou.
\newblock Encoding metal mask projection for metal artifact reduction in computed tomography.
\newblock In \emph{International Conference on Medical Image Computing and Computer-Assisted Intervention}, pages 147--157. Springer, 2020.

\bibitem[Meyer et~al.(2010)Meyer, Raupach, Lell, Schmidt, and Kachelrie{\ss}]{meyer2010normalized}
Esther Meyer, Rainer Raupach, Michael Lell, Bernhard Schmidt, and Marc Kachelrie{\ss}.
\newblock Normalized metal artifact reduction (nmar) in computed tomography.
\newblock \emph{Medical Physics}, 37\penalty0 (10):\penalty0 5482--5493, 2010.

\bibitem[Meyer et~al.(2012)Meyer, Raupach, Lell, Schmidt, and Kachelrie{\ss}]{meyer2012frequency}
Esther Meyer, Rainer Raupach, Michael Lell, Bernhard Schmidt, and Marc Kachelrie{\ss}.
\newblock Frequency split metal artifact reduction (fsmar) in computed tomography.
\newblock \emph{Medical Physics}, 39\penalty0 (4):\penalty0 1904--1916, 2012.

\bibitem[Park et~al.(2015)Park, Hwang, and Seo]{park2015metal}
Hyoung~Suk Park, Dosik Hwang, and Jin~Keun Seo.
\newblock Metal artifact reduction for polychromatic x-ray ct based on a beam-hardening corrector.
\newblock \emph{IEEE Transactions on Medical Imaging}, 35\penalty0 (2):\penalty0 480--487, 2015.

\bibitem[Park et~al.(2017)Park, Chung, Lee, Kim, and Seo]{park2017sinogram}
Hyung~Suk Park, Yong~Eun Chung, Sung~Min Lee, Hwa~Pyung Kim, and Jin~Keun Seo.
\newblock Sinogram-consistency learning in ct for metal artifact reduction.
\newblock \emph{arXiv preprint arXiv:1708.00607}, 1, 2017.

\bibitem[Park et~al.(2018)Park, Lee, Kim, Seo, and Chung]{park2018ct}
Hyoung~Suk Park, Sung~Min Lee, Hwa~Pyung Kim, Jin~Keun Seo, and Yong~Eun Chung.
\newblock Ct sinogram-consistency learning for metal-induced beam hardening correction.
\newblock \emph{Medical Physics}, 45\penalty0 (12):\penalty0 5376--5384, 2018.

\bibitem[Paszke et~al.(2019)Paszke, Gross, Massa, Lerer, Bradbury, Chanan, Killeen, Lin, Gimelshein, Antiga, et~al.]{paszke2019pytorch}
Adam Paszke, Sam Gross, Francisco Massa, Adam Lerer, James Bradbury, Gregory Chanan, Trevor Killeen, Zeming Lin, Natalia Gimelshein, Luca Antiga, et~al.
\newblock Pytorch: An imperative style, high-performance deep learning library.
\newblock \emph{Advances in Neural Information Processing Systems}, 32, 2019.

\bibitem[Ronneberger et~al.(2015)Ronneberger, Fischer, and Brox]{ronneberger2015u}
Olaf Ronneberger, Philipp Fischer, and Thomas Brox.
\newblock U-net: Convolutional networks for biomedical image segmentation.
\newblock In \emph{International Conference on Medical Image Computing and Computer-Assisted Intervention}, pages 234--241. Springer, 2015.

\bibitem[Selles et~al.(2023{\natexlab{a}})Selles, Slotman, van Osch, Nijholt, Wellenberg, Maas, and Boomsma]{selles2023ai}
Mark Selles, Derk~J Slotman, Jochen~AC van Osch, Ingrid~M Nijholt, Ruud~HH Wellenberg, Mario Maas, and Martijn~F Boomsma.
\newblock Is ai the way forward for reducing metal artifacts in ct? development of a generic deep learning-based method and initial evaluation in patients with sacroiliac joint implants.
\newblock \emph{European Journal of Radiology}, 163:\penalty0 110844, 2023{\natexlab{a}}.

\bibitem[Selles et~al.(2023{\natexlab{b}})Selles, van Osch, Maas, Boomsma, and Wellenberg]{selles2023advances}
Mark Selles, Jochen van Osch, Mario Maas, Martijn Boomsma, and Ruud Wellenberg.
\newblock Advances in metal artifact reduction in ct images: A review of traditional and novel metal artifact reduction techniques.
\newblock \emph{European Journal of Radiology}, page 111276, 2023{\natexlab{b}}.

\bibitem[Slavic and Danielsson(2022)]{slavic2022dual}
Scott Slavic and Mats Danielsson.
\newblock Dual-energy: The ge approach.
\newblock In \emph{Spectral Imaging: Dual-Energy, Multi-Energy and Photon-Counting CT}, pages 45--62. Springer, 2022.

\bibitem[Slavic et~al.(2017)Slavic, Madhav, Profio, Crotty, Nett, and Hsieh]{slavic2017gsi}
Scott Slavic, Priti Madhav, Mark Profio, Dominic Crotty, Elizabeth Nett, and Jiang Hsieh.
\newblock Gsi xtream on revolutiontm ct. volume. spectral. simplified.
\newblock 2017.

\bibitem[von Platen et~al.(2022)von Platen, Patil, Lozhkov, Cuenca, Lambert, Rasul, Davaadorj, Nair, Paul, Berman, Xu, Liu, and Wolf]{von-platen-etal-2022-diffusers}
Patrick von Platen, Suraj Patil, Anton Lozhkov, Pedro Cuenca, Nathan Lambert, Kashif Rasul, Mishig Davaadorj, Dhruv Nair, Sayak Paul, William Berman, Yiyi Xu, Steven Liu, and Thomas Wolf.
\newblock Diffusers: State-of-the-art diffusion models.
\newblock \url{https://github.com/huggingface/diffusers}, 2022.

\bibitem[Wang et~al.(1996)Wang, Snyder, O'Sullivan, and Vannier]{wang1996iterative}
Ge Wang, Donald~L Snyder, Joseph~A O'Sullivan, and Michael~W Vannier.
\newblock Iterative deblurring for ct metal artifact reduction.
\newblock \emph{IEEE Transactions on Medical Imaging}, 15\penalty0 (5):\penalty0 657--664, 1996.

\bibitem[Wang et~al.(1999)Wang, Vannier, and Cheng]{wang1999iterative}
Ge Wang, Michael~W Vannier, and Ping-Chin Cheng.
\newblock Iterative x-ray cone-beam tomography for metal artifact reduction and local region reconstruction.
\newblock \emph{Microscopy and Microanalysis}, 5\penalty0 (1):\penalty0 58--65, 1999.

\bibitem[Wang et~al.(2021{\natexlab{a}})Wang, Li, He, Ma, Meng, and Zheng]{wang2021dicdnet}
Hong Wang, Yuexiang Li, Nanjun He, Kai Ma, Deyu Meng, and Yefeng Zheng.
\newblock Dicdnet: deep interpretable convolutional dictionary network for metal artifact reduction in ct images.
\newblock \emph{IEEE Transactions on Medical Imaging}, 41\penalty0 (4):\penalty0 869--880, 2021{\natexlab{a}}.

\bibitem[Wang et~al.(2021{\natexlab{b}})Wang, Li, Zhang, Chen, Ma, Meng, and Zheng]{wang2021indudonet}
Hong Wang, Yuexiang Li, Haimiao Zhang, Jiawei Chen, Kai Ma, Deyu Meng, and Yefeng Zheng.
\newblock Indudonet: An interpretable dual domain network for ct metal artifact reduction.
\newblock In \emph{International Conference on Medical Image Computing and Computer-Assisted Intervention}, pages 107--118. Springer, 2021{\natexlab{b}}.

\bibitem[Wang et~al.(2022{\natexlab{a}})Wang, Li, Meng, and Zheng]{wang2022adaptive}
Hong Wang, Yuexiang Li, Deyu Meng, and Yefeng Zheng.
\newblock Adaptive convolutional dictionary network for ct metal artifact reduction.
\newblock \emph{arXiv preprint arXiv:2205.07471}, 2022{\natexlab{a}}.

\bibitem[Wang et~al.(2022{\natexlab{b}})Wang, Xie, Li, Huang, Meng, and Zheng]{wang2022orientation}
Hong Wang, Qi Xie, Yuexiang Li, Yawen Huang, Deyu Meng, and Yefeng Zheng.
\newblock Orientation-shared convolution representation for ct metal artifact learning.
\newblock In \emph{International Conference on Medical Image Computing and Computer-Assisted Intervention}, pages 665--675. Springer, 2022{\natexlab{b}}.

\bibitem[Wang et~al.(2023{\natexlab{a}})Wang, Li, Zhang, Meng, and Zheng]{wang2023indudonet+}
Hong Wang, Yuexiang Li, Haimiao Zhang, Deyu Meng, and Yefeng Zheng.
\newblock Indudonet+: A deep unfolding dual domain network for metal artifact reduction in ct images.
\newblock \emph{Medical Image Analysis}, 85:\penalty0 102729, 2023{\natexlab{a}}.

\bibitem[Wang et~al.(2023{\natexlab{b}})Wang, Xie, Zeng, Ma, Meng, and Zheng]{wang2023oscnet}
Hong Wang, Qi Xie, Dong Zeng, Jianhua Ma, Deyu Meng, and Yefeng Zheng.
\newblock Oscnet: Orientation-shared convolutional network for ct metal artifact learning.
\newblock \emph{IEEE Transactions on Medical Imaging}, 2023{\natexlab{b}}.

\bibitem[Wang et~al.(2023{\natexlab{c}})Wang, Zhou, Wei, Li, and Zheng]{wang2023mepnet}
Hong Wang, Minghao Zhou, Dong Wei, Yuexiang Li, and Yefeng Zheng.
\newblock Mepnet: A model-driven equivariant proximal network for joint sparse-view reconstruction and metal artifact reduction in ct images.
\newblock In \emph{International Conference on Medical Image Computing and Computer-Assisted Intervention}, pages 109--120. Springer, 2023{\natexlab{c}}.

\bibitem[Wang et~al.(2018)Wang, Zhao, Noble, and Dawant]{wang2018conditional}
Jianing Wang, Yiyuan Zhao, Jack~H Noble, and Benoit~M Dawant.
\newblock Conditional generative adversarial networks for metal artifact reduction in ct images of the ear.
\newblock In \emph{International Conference on Medical Image Computing and Computer-Assisted Intervention}, pages 3--11. Springer, 2018.

\bibitem[Wang et~al.(2024)Wang, Liu, and Li]{wang2024metal}
Yuyang Wang, Xiaomo Liu, and Liang Li.
\newblock Metal artifacts reducing method based on diffusion model using intraoral optical scanning data for dental cone-beam ct.
\newblock \emph{IEEE Transactions on Medical Imaging}, 2024.

\bibitem[Wang et~al.(2004)Wang, Bovik, Sheikh, and Simoncelli]{wang2004image}
Zhou Wang, Alan~C Bovik, Hamid~R Sheikh, and Eero~P Simoncelli.
\newblock Image quality assessment: from error visibility to structural similarity.
\newblock \emph{IEEE Transactions on Image Processing}, 13\penalty0 (4):\penalty0 600--612, 2004.

\bibitem[Watzke and Kalender(2004)]{watzke2004pragmatic}
Oliver Watzke and Willi~A Kalender.
\newblock A pragmatic approach to metal artifact reduction in ct: merging of metal artifact reduced images.
\newblock \emph{European Radiology}, 14:\penalty0 849--856, 2004.

\bibitem[Wu et~al.(2023)Wu, Chen, Wang, Wei, Zhou, Yu, and Zhang]{wu2023unsupervised}
Qing Wu, Lixuan Chen, Ce Wang, Hongjiang Wei, S~Kevin Zhou, Jingyi Yu, and Yuyao Zhang.
\newblock Unsupervised polychromatic neural representation for ct metal artifact reduction.
\newblock \emph{Advances in Neural Information Processing Systems}, 36:\penalty0 69605--69624, 2023.

\bibitem[Yadava et~al.(2014)Yadava, Pal, and Hsieh]{yadava2014reduction}
Girijesh~K Yadava, Debashish Pal, and Jiang Hsieh.
\newblock Reduction of metal artifacts: beam hardening and photon starvation effects.
\newblock In \emph{Medical Imaging 2014: Physics of Medical Imaging}, pages 816--823. SPIE, 2014.

\bibitem[Yu et~al.(2020)Yu, Zhang, Li, and Xing]{yu2020deep}
Lequan Yu, Zhicheng Zhang, Xiaomeng Li, and Lei Xing.
\newblock Deep sinogram completion with image prior for metal artifact reduction in ct images.
\newblock \emph{IEEE Transactions on Medical Imaging}, 40\penalty0 (1):\penalty0 228--238, 2020.

\bibitem[Zhang et~al.(2021)Zhang, Li, Zhang, Zhang, Ma, and Ding]{zhang2021reduction}
Fang-ling Zhang, Ruo-cheng Li, Xiao-ling Zhang, Zhao-hui Zhang, Ling Ma, and Lei Ding.
\newblock Reduction of metal artifacts from knee tumor prostheses on ct images: value of the single energy metal artifact reduction (semar) algorithm.
\newblock \emph{BMC Cancer}, 21:\penalty0 1--9, 2021.

\bibitem[Zhang et~al.(2018)Zhang, Isola, Efros, Shechtman, and Wang]{zhang2018unreasonable}
Richard Zhang, Phillip Isola, Alexei~A Efros, Eli Shechtman, and Oliver Wang.
\newblock The unreasonable effectiveness of deep features as a perceptual metric.
\newblock In \emph{Proceedings of the IEEE conference on Computer Vision and Pattern Recognition}, pages 586--595, 2018.

\bibitem[Zhang and Yu(2018)]{zhang2018convolutional}
Yanbo Zhang and Hengyong Yu.
\newblock Convolutional neural network based metal artifact reduction in x-ray computed tomography.
\newblock \emph{IEEE Transactions on Medical Imaging}, 37\penalty0 (6):\penalty0 1370--1381, 2018.

\bibitem[Zhang et~al.(2010)Zhang, Mou, and Tang]{zhang2010beam}
Yanbo Zhang, Xuanqin Mou, and Shaojie Tang.
\newblock Beam hardening correction for fan-beam ct imaging with multiple materials.
\newblock In \emph{IEEE Nuclear Science Symposuim \& Medical Imaging Conference}, pages 3566--3570. IEEE, 2010.

\bibitem[Zhu et~al.(2021)Zhu, Han, Xi, Li, and Yan]{zhu2021completion}
Linlin Zhu, Yu Han, Xiaoqi Xi, Lei Li, and Bin Yan.
\newblock Completion of metal-damaged traces based on deep learning in sinogram domain for metal artifacts reduction in ct images.
\newblock \emph{Sensors}, 21\penalty0 (24):\penalty0 8164, 2021.

\end{thebibliography}
}


\end{document}